\documentclass[letterpaper]{article} 
\usepackage{aaai2026}  
\usepackage{times}  
\usepackage{helvet}  
\usepackage{courier}  
\usepackage[hyphens]{url}  
\usepackage{graphicx} 
\usepackage{natbib}  
\usepackage{caption} 
\usepackage{algorithm}
\usepackage{algorithmic}

\usepackage{newfloat}
\usepackage{listings}
\DeclareCaptionStyle{ruled}{labelfont=normalfont,labelsep=colon,strut=off} 
\floatstyle{ruled}
\newfloat{listing}{tb}{lst}{}
\floatname{listing}{Listing}
\usepackage{multirow,amsmath}
\usepackage{amsfonts} 

\title{HandMCM: Multi-modal Point Cloud-based Correspondence State Space Model for 3D Hand Pose Estimation}
\author{
    Wencan Cheng, Gim Hee Lee\\
}
\affiliations{
    Department of Computer Science, National University of Singapore \\
    \{wccheng,  gimhee.lee\}@nus.edu.sg
}

\usepackage{bibentry}

\begin{document}

\maketitle
\begin{abstract}


3D hand pose estimation that involves accurate estimation of 3D human hand keypoint locations is crucial for many human-computer interaction applications such as augmented reality. However, this task poses significant challenges due to self-occlusion of the hands and occlusions caused by interactions with objects. In this paper, we propose HandMCM to address these challenges. Our HandMCM is a novel method based on the powerful state space model (Mamba). By incorporating modules for local information injection/filtering and correspondence modeling, the proposed correspondence Mamba effectively learns the highly dynamic kinematic topology of keypoints across various occlusion scenarios. Moreover, by integrating multi-modal image features, we enhance the robustness and representational capacity of the input, leading to more accurate hand pose estimation. Empirical evaluations on three benchmark datasets demonstrate that our model significantly outperforms current state-of-the-art methods, particularly in challenging scenarios involving severe occlusions. These results highlight the potential of our approach to advance the accuracy and reliability of 3D hand pose estimation in practical applications. 

\end{abstract}
    
\section{Introduction}


3D hand pose estimation (HPE) aims to determine the 3D locations of hand keypoints from sensor-acquired visual data. As it provides a fundamental understanding of hand actions and behaviors, HPE plays a crucial role in human-computer interaction applications such as augmented/virtual reality, robotics, \emph{etc}.. However, severe self-occlusions and hand-object occlusions during hand-object interactions make HPE a challenging task.

\begin{figure}[!h]
\centering
\includegraphics[width=.9\linewidth]{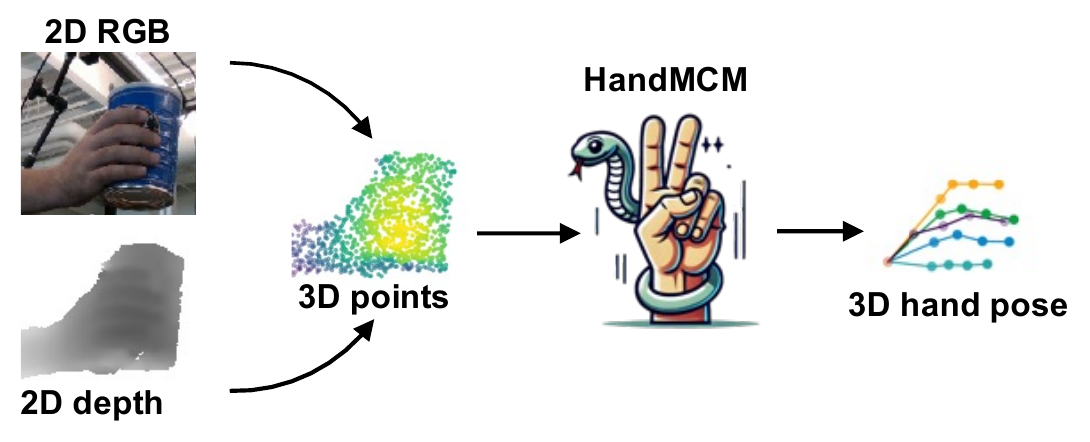}
\caption{
Illustration of our HandMCM concept. The model extracts multi-modal 3D point features from input depth images, RGB images, and corresponding point clouds. The Mamba architecture directly explores geometric and kinematic correspondences in 3D point space to accurately recover hand poses.}
\label{fig:concept}
\end{figure}


With the emergence of deep learning technologies, early deep learning-based methods utilized conventional convolutional neural networks (CNNs) to estimate hand poses from 2D images \cite{tompson2014real, ge2016robust,  ren2019srn, chen2020pose, fang2020jgr, ren2021pose, du2019crossinfonet}. To address the challenging non-linear mapping problem from 2D images to 3D poses, CNN architectures that leverage 3D modalities such as point clouds \cite{ge2018hand, ge2018point, cheng2021handfoldingnet, cheng2023handr2n2}, voxelized representations \cite{ge20173d, moon2018v2v}, or multi-modal combinations \cite{ren2023two} have been extensively studied. 


Despite the notable progress made by these approaches, accurate estimation remains challenging in scenarios involving hand-object occlusion and self-occlusion. To facilitate the reconstruction of occluded keypoints using information from visible keypoints, recent studies \cite{cheng2023handr2n2, ren2023two, wang2023handgcnformer}, such as Hamba \cite{dong2024hamba}, have attempted to capture the spatial kinematic relationships between keypoints through graph convolution. However, due to the wide variety of hand poses and interactions, those static graph-based methods struggle to accurately model complex occlusions and the dynamic spatial relationships between keypoints.


To address these challenges, we propose a novel architecture named \textit{\textbf{M}ulti-modal \textbf{C}orrespondence \textbf{M}amba} (HandMCM), which is the first to implement a state space model to capture dynamic kinematic correspondences for accurate 3D hand pose estimation under various occlusion conditions and complex poses. The proposed Correspondence Mamba mechanism demonstrates superiority in modeling dynamic kinematic correspondence compared to the straightforward combination of an auxiliary graph mechanism in Hamba \cite{dong2024hamba}. \textit{This advantage stems from the intuitive motivation that spatial correspondence can be effectively processed as  sequential data \cite{liu2017retrosynthetic} which is highly compatible with the Mamba model.}
To further enhance accuracy, HandMCM incorporates a local geometric injection and filtering mechanism that explicitly provides the Mamba module with local geometric information such as visible keypoints and occluded regions. This enables the Mamba mechanism to automatically activate kinematically corresponding features for occluded/visible keypoint estimation according to local details.
Additionally, HandMCM employs a powerful multi-modal super point extraction approach that enhances the representation ability of 3D point clouds by incorporating visual cues such as occlusions and interacting objects from color and depth images. 


We evaluate HandMCM on four challenging benchmarks, including the single-hand NYU dataset \cite{tompson2014real}, and the hand-object DexYCB \cite{chao2021dexycb} and HO3D \cite{hampali2020honnotate} datasets. The results show that our framework achieves state-of-the-art performance with a mean distance error of 7.06mm on the NYU dataset. Additionally, our model achieves state-of-the-art results on the hand-object DexYCB and HO3D datasets with mean distance errors of 6.67mm and 1.71cm, respectively. Our source code is publicly available at https://github.com/cwc1260/HandMCM.git.

Our \textbf{main contributions} are summarized as follows:
\begin{itemize}
\item 
We propose HandMCM, a novel Mamba-based architecture for accurate 3D hand pose estimation using RGBD images as multi-modal inputs.
\item We propose a novel correspondence state space model designed to capture the kinematic relationships between keypoints, ensuring high reliability under occlusion.
\item We conduct comprehensive experiments on large and challenging benchmarks, demonstrating the state-of-the-art performance of our proposed method.
\end{itemize}

\section{Related Work}
\subsection{3D Hand Pose Estimation}

Conventional methods utilizing 2D convolutional neural networks (CNNs) \cite{tompson2014real, ge2016robust,  ren2019srn, chen2020pose, fang2020jgr, ren2021pose, du2019crossinfonet} have become popular for 3D HPE due to their ease of implementation. However, these approaches have significant drawbacks in their inability to fully capture the complexities of 3D structures and their dependence on camera perspective.

To address these limitations, methodologies based on 3D CNNs \cite{ge20173d, moon2018v2v} have been developed that use 3D voxel grids derived from depth images to represent volumetric data.
Although these 3D techniques have shown progress in hand pose estimation, they require considerable memory and computational power, which can limit their practicality in real-world applications.

On the other hand, approaches based on PointNet \cite{ge2018hand, ge2018point, chen2018shpr, li2019point, cheng2021handfoldingnet} analyze point cloud data, enabling an accurate representation of 3D structures. PointNet is a deep learning architecture designed to process irregular and unstructured point clouds, and it has been first applied to hand pose estimation in HandPointNet \cite{ge2018hand}.
Subsequent improvements, such as the Point-to-Point model \cite{ge2018point} and SHPR-Net \cite{chen2018shpr} have improved performance by producing point-wise probability distributions. In particular, SHPR-Net \cite{chen2018shpr} integrated HandPointNet with an additional semantic segmentation network to improve performance. More recently, HandFoldingNet \cite{cheng2021handfoldingnet} introduces a novel folding approach to transform a predefined 2D hand skeleton into various hand poses, thus improving accuracy in estimation. Following this approach, HandR2N2 \cite{cheng2023handr2n2} proposes a folding-based recurrent framework to iteratively refine the locations of keypoints through local point region searches. However, a significant limitation of working with point clouds is the high computational cost associated with neighbor querying within dense point sets during convolution operations. As a result, many existing methods typically resort to utilizing sparse point clouds, which can hinder overall performance.

To bridge the shortcomings of images and point clouds, IPNet \cite{ren2023two} and Keypoint-Fusion \cite{liu2024keypoint} suggest the utilization of multi-modal representations, one combining 2D depth images and 3D point clouds while the other combining depth images and RGB images. As a result, the model adeptly extracts dense detail information while effectively capturing 3D spatial features, thereby facilitating accurate 3D hand pose estimation. 


\subsection{State Space Models}
Classical state space models are originally developed in the context of linear dynamic systems in control theory and popularized by Kalman filtering \cite{kalman1960new}, where the state variables and their transitions are described through first-order differential equations. Recently, a new series of state space models \cite{gu2020hippo, gu2021combining, gu2022parameterization, gu2023mamba} has emerged in natural language processing, leveraging their unique interpretability as a bridge between recurrent neural networks and convolutional neural networks to accelerate the learning of a long sequential context. 

The most representative Structured State Space Sequence (S4) model \cite{gu2021combining, gu2022parameterization}, 
demonstrates its superior ability to effectively model dependencies within sequential data. Subsequently, the Mamba framework \cite{gu2023mamba} improves the S4 models by incorporating a linearly scalable projection matrix that adapts to the length of the input sequence for long-sequence models. Additionally, it includes a selection mechanism to selectively propagate or forget information along the sequence or scan path based on the current token. Although initially applied in natural language processing, Mamba-based architectures have recently been explored in computer vision tasks. With their global receptive field and dynamic weight mechanisms, these architectures show promise in image classification \cite{liu2024vmamba, zhu2024vision}, video understanding \cite{li2025videomamba}, and motion generation \cite{zhang2025motion}. Moreover, Mamba-based architectures have also been investigated in other modalities such as 3D point clouds \cite{liang2024pointmamba, liu2024point} and graph representations \cite{wang2024graph}, enabling efficient 3D vision recognition and robust long-range graph prediction.

The research most closely aligned with ours involves the application of the Mamba hand reconstruction, Hamba \cite{dong2024hamba}. However, Hamba employs graph convolution for correspondence extraction, which fails to fully exploit the potential of the Mamba architecture. In contrast, our approach focuses on harnessing the capability of Mamba to model dynamic interaction patterns among keypoints. Furthermore, Hamba neglects the incorporation of local geometry, which, as demonstrated by our study, plays a critical role in improving the accuracy of pose estimation. 

\section{Preliminaries}
\subsection{State Space Model}

A state space model (SSM) originated from the classical 
linear control theory is used to describe the dynamic relationship between a continuous-time scalar input \( u(t) \) and output \( y(t) \) through the following set of differential equations:
\begin{equation}
x'(t) = \boldsymbol{A} x(t) + \boldsymbol{B} u(t), \quad 
y(t) = \boldsymbol{C} x(t) + \boldsymbol{D} u(t),  
\end{equation}
where \( x(t) \in \mathbb{R}^{N} \) represents the continuous-time state vector, while \( x'(t) \) is its derivative. \( \boldsymbol{A} \in \mathbb{R}^{N \times N} \), \( \boldsymbol{B} \in \mathbb{R}^{N \times 1} \), \( \boldsymbol{C} \in \mathbb{R}^{1 \times N} \), and \( \boldsymbol{D} \in \mathbb{R}^{1 \times 1} \) are weighting parameters.

\subsection{Discretization of SSM}

The continuous SSM must be discretized to be deployed in deep learning models that generally accept discrete inputs.
Let a discrete sequence of scalar inputs be \( u_1, \ldots, u_L \), the SSM can be reformulated into a discrete-time version with the following recursive expressions:
\begin{equation}
    x_k = \boldsymbol{\overline{A}} x_{k-1} + \boldsymbol{\overline{B}} u_k, \quad 
    y_k = \boldsymbol{\overline{C}} x_k + \boldsymbol{\overline{D}} u_k,
\end{equation}
where \( \boldsymbol{\overline{A}}, \boldsymbol{\overline{B}}, \boldsymbol{\overline{C}}, \boldsymbol{\overline{D}} \) represents transformations of the original parameters according to a specific discretization step.

\subsection{Parallelization of SSM}
The recursive relations make the discrete-time SSM resemble the structure of an RNN, where \( x_k \in \mathbb{R}^N \) plays the role of a hidden state at time step \( k \). More importantly, due to the linear structure of the recursion, it is possible to compute the entire output sequence \( y_1, \ldots, y_L \) directly by applying a convolution with a fixed kernel \( \boldsymbol{\overline{K}} \in \mathbb{R}^{L} \):
\begin{equation}
\begin{aligned}
    \boldsymbol{\overline{K}} & = (\boldsymbol{\overline{C}}\boldsymbol{\overline{B}}, \boldsymbol{\overline{C}}\boldsymbol{\overline{A}}\boldsymbol{\overline{B}}, \dots, \boldsymbol{\overline{C}}\boldsymbol{\overline{A}}^{L-1}\boldsymbol{\overline{B}}), \\
    y & = \boldsymbol{\overline{K}} \ast u.
\end{aligned}
\end{equation}
Once trained, the kernel \( \boldsymbol{\overline{K}} \) serves as a comprehensive representation of the SSM, simplifying the model to a one-dimensional convolution with an extended kernel length.
\section{
Our Method}
\begin{figure*}[!t]
\setlength{\abovecaptionskip}{0cm}
\centering
\includegraphics[width=0.8\linewidth]{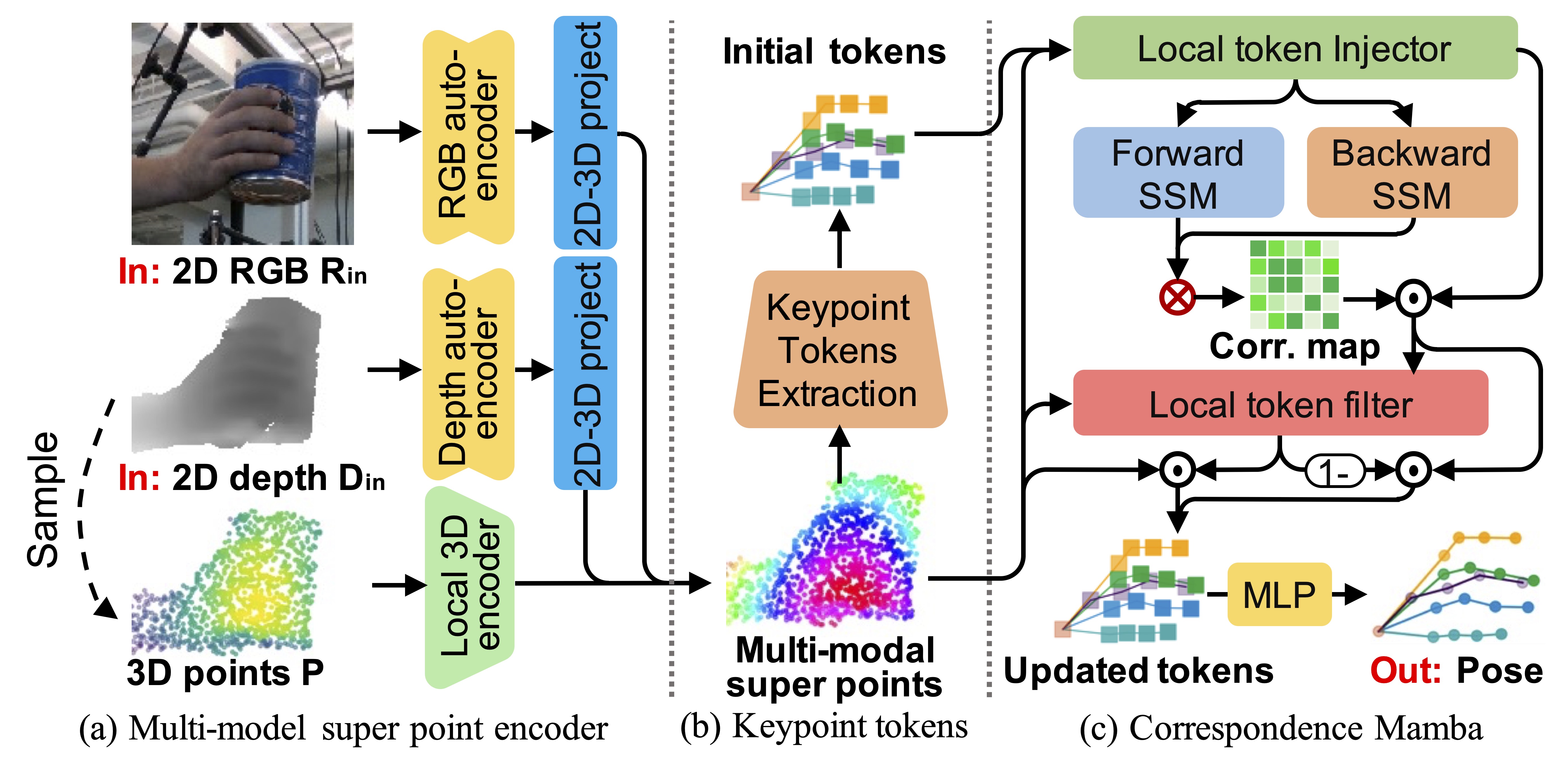}
\caption{The HandMCM architecture. HandMCM takes as input an RGB image, a 2D depth image. The multi-modal super point encoder utilizes a PointNet-based local 3D encoder and two 2D autoencoders to extract local 3D point features and local 2D features from the RGB and depth inputs, respectively. The encoder then aggregates these projected 2D features with the 3D features, forming multi-modal super point features. 
Keypoint tokens are derived from the super points and fed into the novel correspondence Mamba to accurately estimate 3D keypoint coordinates.}
\label{fig:architecture}
\end{figure*}

The architecture of our HandMCM is illustrated in Figure~\ref{fig:architecture}. Formally, 
the architecture of our HandMCM accepts multi-modal inputs 
which includes a hand depth image $\mathbf{D}_{in} \in \mathbb{R}^{H \times W}$ 
and an RGB image $\mathbf{R}_{in} \in \mathbb{R}^{H \times W \times 3}$. It also randomly samples a set of pixels from $\mathbf{D}_{in}$ and projects them into 3D point coordinates $\mathbf{P} \in \mathbb{R}^{N \times 3}$ to form the additional 3D perceptivity. The architecture produces output in the form of 3D keypoint coordinates $\mathbf{J} \in \mathbb{R}^{J \times 3}$. 

Initially, HandMCM samples a multi-modal super point set that integrates features from input RGB and depth images, thereby providing rich visual information including occlusion cues for subsequent processing stages. From this super point set, keypoint tokens are extracted and used as input to the Mamba block. The proposed novel correspondence Mamba block iteratively refines these keypoint tokens by leveraging the intrinsic kinematic relationships among keypoints and their local features to achieve an accurate estimation of keypoint coordinates.

\subsection{Multi-Modal Super Point Encoder}

Given that the input point set is irregular and unordered, our HandMCM utilizes a point set convolutional layer \cite{qi2017pointnet++, liu2019flownet3d} as a local 3D encoder. This layer extracts a subsampled 3D super point set $\mathbf{P'} \in \mathbb{R}^{N/2 \times 3}$ alongside the corresponding 3D local geometric features $\mathbf{F}_{p} \in \mathbb{R}^{N/2 \times C_{p}}$. 

Subsequently, the depth and RGB images are each processed by independent ResNet-based autoencoders, resulting in 2D local visual feature maps $\mathbf{F}_{d} \in \mathbb{R}^{H/2 \times W/2 \times C_{d}}$ and $\mathbf{F}_{rgb} \in \mathbb{R}^{H/2 \times W/2 \times C_{rgb}}$, respectively. These 2D local visual features undergo 2D-3D projection and interpolation to create 3D point features $\mathbf{F}_{d \rightarrow p} \in \mathbb{R}^{N/2 \times C_{d}}$ and $\mathbf{F}_{rgb \rightarrow p} \in \mathbb{R}^{N/2 \times C_{rgb}}$. Following this projection, the features are concatenated with the super points, resulting in a combined feature set $\mathbf{F} \in \mathbb{R}^{N/2 \times (C_{p}+C_{d}+C_{rgb})}$. Finally, the super points are input into the keypoint token extraction module, which generates initial keypoint tokens for the Mamba blocks.

\subsection{Keypoint Token Extraction}
In our HandMCM framework, the keypoint tokens $\textbf{X} \in \mathbb{R}^{J \times C}$, where the length is the number of tokens $J$, and the token size is $C$, serve as the starting point for processing within the proposed Mamba block. To construct the initial keypoint tokens, we utilize the joint-wise embeddings initially introduced by HandFoldingNet \cite{cheng2021handfoldingnet}.

The keypoint tokens initialization module processes the super points, incorporating their features $\mathbf{F}$ through two point set convolutional layers to generate 3D global vectors. The initial keypoint tokens (embeddings) $\textbf{X}_0$ are then performed sequentially using a three-layer bias-induced layer (BIL) \cite{cheng2019point}. In this process, the global vector is replicated $J$ times and fed into the BIL layers. This results in the generation of embeddings for the $J$ individual keypoints. Importantly, the BIL provides keypoint-wise independent biases, which function similarly to learnable positional embeddings, which enhance the flexibility and accuracy of the keypoint representation.
After the acquisition of the keypoint tokens, a linear transformation is applied to project the high-dimensional tokens into the 3D space as the positions $\textbf{J}_0$ for the local feature collection.

\subsection{Correspondence Mamba}

The correspondence Mamba illustrated in Figure \ref{fig:architecture} is the key contributing component of our proposed framework. It can effectively model the dynamic kinematic topology of keypoints by building a robust correspondence map by a state space model.

\subsubsection{Bi-directional Correspondence SSM}

The basic architecture follows the bidirectional gated SSM (BiGS) \cite{wang2022pretraining}. Specifically, let $\mathbf{X}_{i-1} \in \mathbb{R}^{
J\times C}$ be keypoints tokens for the $k$-th Mamba block. The Mamba block first transforms tokens to construct the bidirectional inputs $\mathbf{X}_f$ and $\mathbf{X}_b$, \emph{i.e.}:
\begin{equation}
\begin{aligned}
    &\widetilde{\mathbf{X}}=\operatorname{LN}(\mathbf{X}_{k-1} ),~
\mathbf{V}=\operatorname{GELU}(\mathbf{W}_v \widetilde{\mathbf{X}}),\\
&\mathbf{X}_f=\operatorname{GELU}(\mathbf{W}_{f} \widetilde{\mathbf{X}}),~ 
\mathbf{X}_b=\operatorname{GELU}(\mathbf{W}_{b} \operatorname{Reverse}(\widetilde{\mathbf{X}})),   
\end{aligned}
\label{eq.norm}
\end{equation}
where 
$\operatorname{LN}(\cdot)$, $\operatorname{GELU}(\cdot)$ and $\operatorname{Reverse}(\cdot)$ are the layer normalization \cite{ba2016layer}, Gaussian error linear unit, and order flip operation on the token length dimension, respectively.
$\mathbf{W}_v, \mathbf{W}_{f}$ and $\mathbf{W}_{b}$ denote the learnable linear weights.

Subsequently, the bidirectional inputs $\mathbf{X}_f$ and $\mathbf{X}_b$ are sent to two sequential blocks of forward and backward SSM layers to model the correspondence between keypoints:
\begin{equation}
\mathbf{U}_f = \mathbf{W}_{u_f} \text{SSM}(\mathbf{X}_f), \quad
\mathbf{U}_b = \mathbf{W}_{u_b} \text{SSM}(\mathbf{X}_b),
\end{equation}
where $\mathbf{W}_{u_f}$ and $\mathbf{W}_{u_b}$ are learnable parameter matrices for each direction.
Based on the bidirectional 
correspondence between keypoints $\mathbf{U}_f$ and $\mathbf{U}_b$, we explicitly construct a dynamic correspondence map that fuses the correlation of each pair of keypoints by an outer product \cite{he2018outer}:
\begin{equation}
\mathbf{M}_{corr} = \mathbf{W}_c (\mathbf{U}_f \otimes \operatorname{Reverse}(\mathbf{U}_b)),
\end{equation}
where $\mathbf{W}_c$ is the trainable transformation weights. As shown in Figure \ref{fig:architecture} (c), each element in the correspondence map contains both forward and backward information.
Finally, the correlation map is multiplied with the transformed token $\mathbf{V}$ to obtain an updated token:
\begin{equation}
\mathbf{X}_k = \mathbf{M}_{corr}  \mathbf{V}.
\label{eq.update}
\end{equation}

For the pose regression, a transformation weight $\mathbf{W}_r \in \mathbb{R}^{C \times 3}$ is learned to transform the updated high-dimensional tokens into the 3D keypoint coordinates,
\begin{equation}
\mathbf{J}_k =\mathbf{X}_k\mathbf{W}_r  .
\label{eq.reg}
\end{equation}

\subsubsection{Local Token Injection \& Filtering}
Although the suggested correspondence Mamba can capture dynamic kinematics, 
the performance remains unsatisfactory since the local geometric information around the keypoints is ignored.
To address this issue, we propose a unique local token injection and filtering mechanism to provide the Mamba block with local information and to tailor the output appropriately.

Following previous studies \cite{cheng2023handr2n2, cheng2024handdiff}, we first group the local super point sets $\mathcal{N}_{\mathbf{j}_j}$ with corresponding features near each input keypoint $\mathbf{j}_j \in \mathbf{J}_k$ using the k-nearest neighbor algorithm. Each feature $\mathbf{f} \in \mathbf{F}$ is then concatenated with its token $\mathbf{x}_j \in \mathbf{X}$ and its local coordinate $\mathbf{p} - \mathbf{j}_j$. 
Subsequently, the local token is computed by the point set convolution operation \cite{qi2017pointnet, liu2019flownet3d}, which aggregates local geometry, as follows: 
\begin{equation}
\mathbf{x}_{loc,i} = \mathop{\operatorname{SetConv}}\limits_{(\mathbf{p}, \mathbf{f}) \in \mathcal{N}_{\mathbf{j}_j}} ([\mathbf{p} - \mathbf{j}_j, \mathbf{x}_j, \mathbf{f}]).
\end{equation}
Once the local tokens are prepared, we inject them into the Mamba block. To achieve this, we reformulate the layer normalization $\operatorname{LN}(\cdot)$ from Eq.~\eqref{eq.norm} by multiplying the local token with the input as follows,
\begin{equation}
\widetilde{\mathbf{X}}=\operatorname{LN}(\mathbf{X}_{k-1} \odot \mathbf{X}_{loc}).\\
\end{equation}
%
For output filtering, the local geometry is first aggregated from the concatenation of local point sets and updated tokens from Eq.~\eqref{eq.update}. 
The filter coefficients are then calculated on local features by applying the sigmoid activation:
\begin{equation}
\mathbf{G} = \sigma(\mathbf{X}_{loc}).
\end{equation}
Finally, the updated tokens are filtered along with the local geometry. 
We can thus reformulate 
Eq.~\eqref{eq.reg} for the final keypoint coordinate regression as the weighted sum of updated correspondence tokens and local tokens given by:
\begin{equation}
\mathbf{J}_k =  ( \mathbf{G} \odot \mathbf{X}_k+ (1-\mathbf{G}) \odot \mathbf{X}_{loc} ) \mathbf{W}_r,
\end{equation}
where $\mathbf{W}_r$ is the same transformation weight in Eq. \eqref{eq.reg}.

\subsection{Loss Function}

Since both the keypoint token initializer and Mamba blocks generate their respective estimated coordinates, all outputs are supervised through a loss function as follows:
\begin{equation}
\mathcal L = \sum_{k=0}^K \sum_{j=1}^J L1_{smooth}(\textbf{j}_{k,j} - \textbf{j}_j^*),
\end{equation}
where $ \textbf{j}_j^*$ indicates the ground-truth coordinate of the $j$-th joint, and $K$ indicates the quantity of stacked Mamba blocks.
As suggested in previous regression studies \cite{ren2019srn, cheng2021handfoldingnet}, we employ the smooth $L1$ function because it is less sensitive to outliers than the $L2$ function. The smooth $L1$ function is defined as follows:
\begin{equation}
	L1_{smooth}(\textbf{x}) = \begin{cases}
	0.5|\textbf{x}|, &|\textbf{x}|<0.01\\
	|\textbf{x}|-0.005, &\text{otherwise}
		   \end{cases}
    .
\end{equation}

\section{Experiments}
\subsection{Datasets and Evaluation Metrics}

\noindent
{\bf NYU} \cite{tompson2014real} is a dataset for \textit{single hand} pose estimation, which provides depth images of single hands captured from three distinct views using the PrimeSense 3D sensor. It contains 72K training frames and 8K testing frames. 
Following prior research practices \cite{ge2018point, cheng2021handfoldingnet}, 
14 keypoints from the complete set of 36 annotated joints and a single view are used for evaluation.
This strategy aligns with previous work to maintain standardized comparisons across studies while ensuring reliable benchmarking.

\noindent
{\bf DexYCB} \cite{chao2021dexycb} is a benchmark for \textit{hand-object} interactions, containing 582K frames annotated with 21 hand keypoints from eight camera views. It involves 10 subjects interacting with 20 standardized YCB objects and includes four evaluation protocols: S0 (seen subjects, views, and objects), S1 (unseen subjects), S2 (unseen views) and S3 (unseen objects). These protocols and standardized objects make DexYCB a robust resource for testing models in various real-world scenarios.

\noindent
{\bf HO3D\_v2} \cite{hampali2020honnotate} is another benchmark for \textit{hand-object} interactions, including 66K training images and 11K testing images with 10 subjects and 10 objects. It provides detailed 3D pose annotations for hands, with test results evaluated through an online submission system specific to this dataset.

\noindent
{\bf Evaluation Metrics.} To assess the performance of HPE, we utilize a widely adopted metric: the mean keypoint error (MKE). The mean keypoint error measures the average Euclidean distance between the predicted and ground-truth keypoint positions, indicating model precision. 

\subsection{Implementation Details}
We 
implement the model using PyTorch on a NVIDIA 4090 GPU and 
train it with AdamW optimizer \cite{loshchilov2017decoupled} (beta$_1$ = 0.5, beta$_2$ = 0.999) and a learning rate of 0.001. The input depth and RGB images 
are resized to 128$\times$128 pixels, and the input point clouds 
are downsampled to 1,024 points from the input depth. We 
use a batch size of 32 for training. For data augmentation, we 
apply random rotations ranging from -180 to 180 degrees, 3D scaling within a range of 0.9 to 1.1, and 3D translations between -10 and 10 mm. In addition, we set the number of super points to 256. We 
train the model on the HO3D dataset for 24 epochs, applying learning rate decay at the 19th epoch. For the DexYCB dataset, we 
train for 20 epochs with decay at the 15th epoch. 
For the NYU dataset, training extended over 40 epochs 
with learning rate decay proceeding at the 30th epoch. 
Note that we bypass the RGB data for the training and evaluation on NYU
due to its low-quality RGB data which has massive unpredictable holes and artifacts.
\begin{figure*}
\centering
\includegraphics[width=.95\linewidth]{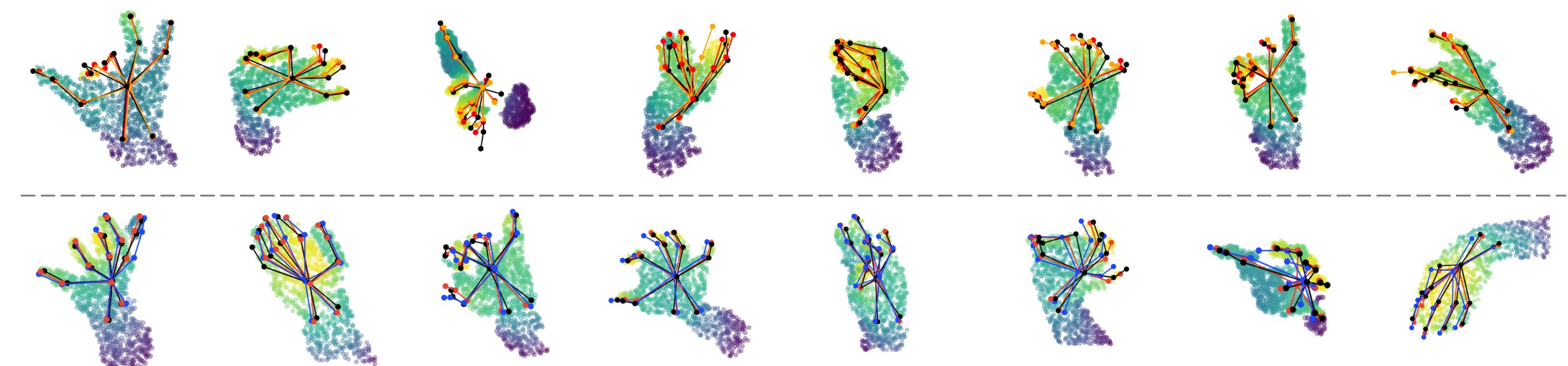}
\caption{Qualitative results of the HandMCM model on the NYU dataset. 
3D points are 
colored in the figure to distinctly illustrate occlusions. The ground truth joint coordinates are displayed in black, the results of the previous state-of-the-art model, HandDAGT \cite{cheng2024handdagt}, are shown in orange (top), the results of Hamba \cite{dong2024hamba} are shown in blue (bottom), and the estimated joint coordinates produced by our model are depicted in red. Best view in color.}
\label{fig:vis}

\end{figure*}

\subsection{Comparison with State-of-the-Art Methods}

\noindent
{\bf NYU Dataset.} We 
conduct a comparison of our method with various state-of-the-art techniques using the single hand NYU dataset \cite{oberweger2017deepprior++, wan2018dense, chen2020pose, cheng2024handdagt, du2019crossinfonet, fang2020jgr, ren2021spatial, ren2021pose, ge20173d, chen2018shpr, ge2018hand, ge2018point, moon2018v2v, cheng2021handfoldingnet, cheng2023handr2n2, ren2023two, liu2024keypoint, cheng2024handdagt}.

\begin{table}[t] \small
\centering
\tabcolsep=2pt
\begin{tabular}{l|c|c}
\hline
 Method & MKE (mm)& Input \\
\hline
DeepPrior++ (Oberweger et al. 2017)    & 12.24         &Depth  \\
Pose-Ren \cite{chen2020pose}               & 11.81         &Depth  \\
SHPR-Net \cite{chen2018shpr}               & 10.78         &Point  \\
HandPointNet \cite{ge2018hand}               & 10.54         &Point  \\
DenseReg \cite{wan2018dense}               & 10.20          &Depth  \\
CrossInfoNet \cite{du2019crossinfonet}        & 10.08         &Depth  \\
Point-to-Point \cite{ge2018point}            & 9.10          &Point  \\
HandFolding (Cheng et al. 2021)       & 8.58       &Point  \\
V2V (Moon et al. 2018)                    & 8.42          &Voxel  \\
Hamba$^\dag$ \cite{dong2024hamba} & 8.38 & Depth \\
JGR-P2O \cite{fang2020jgr}                  & 8.29          &Depth  \\
HandGCNFormer \cite{wang2023handgcnformer}      & 7.43       &Point  \\
SSRN \cite{ren2021spatial}                & 7.37          &Depth  \\
HandDiff \cite{cheng2024handdiff} & 7.38  & Depth  \\
PHG \cite{ren2021pose}                    & 7.39          &Depth  \\
HandR2N2 \cite{cheng2023handr2n2}  & 7.27 & Point \\
IPNet \cite{ren2023two}   & 7.17  &Depth  \\
HandDAGT \cite{cheng2024handdagt} & 7.12 & Depth \\
\hline
 HandMCM (Ours)  & \textbf{7.06}  & Depth\\
\hline
\end{tabular}
\caption{Comparison of the proposed method with previous state-of-the-art methods on the NYU datasets. Note, Voxels \& Points are transformed and sampled from Depth.} 
\label{tab:NYU}
\end{table}

The results of our evaluations measured by the mean joint error on the NYU dataset are summarized in Table \ref{tab:NYU}. Our proposed 
HandMCM achieves a mean joint error of 7.06 mm, indicating its performance at the state-of-the-art level. Furthermore, Figure \ref{fig:vis} showcases qualitative results of HandMCM on the NYU dataset, illustrating its effectiveness in accurately estimating hand keypoints, even in scenarios involving occlusions.

\begin{figure*}
\setlength{\abovecaptionskip}{0cm}
\centering
\includegraphics[width=\linewidth]{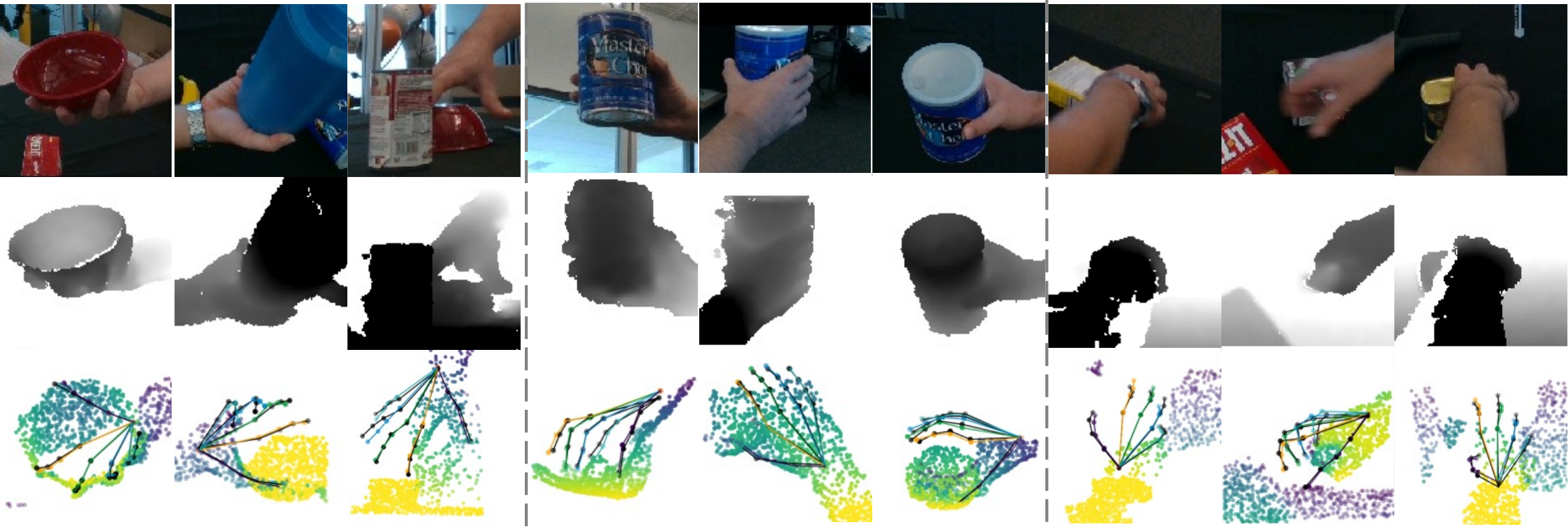}

\caption{Qualitative results of HandMCM on the DexYCB datasets illustrating different object occlusions (left), view-angle-dependent occlusions (middle) and self-occlusions (right). Input 3D points and output keypoint coordinates are depicted within the same frame (third rows) and rotated at top-down view angles in order to clearly \textit{view occlusions from above} as shown in the figure. Ground truth is shown in black and the estimated joint coordinates of our model are shown in colors. Best view in color.}
\label{fig:dex}

\end{figure*}

\noindent
{\bf DexYCB Dataset.}
We assess our HandMCM on the hand-object dataset DexYCB and compare its performance against several existing state-of-the-art methods \cite{xiong2019a2j, spurr2020weakly, lin2021end, tse2022collaborative, park2022handoccnet, ren2023two, cheng2024handdagt, liu2024keypoint, cheng2024handdiff}, adhering to the official dataset split protocols. As indicated in Table \ref{tab:dexycb}, our method outperforms the latest state-of-the-art (SOTA) techniques across all four evaluation protocols. 
The results highlight the exceptional capability of our HandMCM in managing occlusions, as it significantly exceeds other models by 11.5\% on the DexYCB dataset. Furthermore, the qualitative results presented in Figure \ref{fig:dex} demonstrate that HandMCM effectively estimates poses during hand-object interactions, even in the presence of various occlusions.

\noindent
{\bf HO3D Dataset.}
Additionally, we benchmark the method on the HO3D dataset against leading methods \cite{li2021hybrik,yang2022artiboost,park2022handoccnet,malik2021handvoxnet++, ren2023two, cheng2024handdagt, liu2024keypoint}, with the state-of-the-art results shown in Table \ref{tab:ho3d}. Our proposed HandMCM achieves the mean keypoint error of 1.71 cm. It also surpasses Hamba \cite{dong2024hamba}. These findings demonstrate that our model is highly robust and accurate in scenarios with dynamic occlusions.

\noindent
{\bf Runtime cost. } Our model achieves an average runtime of 15.4 ms per frame on the input RGBD modality with 0.3ms of point subsampling, when tested on an NVIDIA RTX 4090 GPU. It runs faster than the recent SOTA K-Fusion, which requires 17.8 ms on the same GPU.

\begin{table}[t!] \small
\centering
\tabcolsep=2pt
\begin{tabular}{l|cccc|c|c}
\hline
 \multirow{2}{*}{Method} & \multicolumn{5}{c|}{Mean Keypoint Error (mm)}& \multirow{2}{*}{Input}\\
\cline{2-6}
            & S0    &S1     & S2    & S3    & AVG   &  \\
\hline
A2J      & 23.93 & 25.57 & 27.65 & 24.92 & 25.52 & Depth\\
Spurr et al.  & 17.34 & 22.26 & 25.49 & 18.44 & 18.44 & RGB\\
METRO     & 15.24 &-&-&-&-& RGB \\
Tse et al.  & 16.05 & 21.22 & 27.01 & 17.93 & 20.55 & RGB \\
HandOccNet   & 14.04 &-&-&-&-& RGB \\
IPNet       & 8.03  & 9.01  & 8.60  & 7.80  & 8.36  & Depth\\
HandDiff & 7.66  & 8.73  & 8.40  & 7.53 & 8.07 & Depth \\
HandDAGT        &7.72  & 8.68  & 8.22  & 7.52 & 8.03  & Depth \\
K-Fusion  & 6.94  & 8.64  &  7.56  &  7.02  &  7.54  & RGBD \\
\hline
HandMCM (Ours) & \textbf{6.27}  & \textbf{7.86}  & \textbf{6.20}  & \textbf{6.30} & \textbf{6.67}  & RGBD \\
\hline
\end{tabular}
\caption{Comparison of the proposed method with previous state-of-the-art methods on the DexYCB datasets.} 
\label{tab:dexycb}
\end{table}

\begin{table}[t!] \small
\centering
\tabcolsep=2pt
\begin{tabular}{l|c|c}
\hline
Method & MKE (cm)& Input\\
\hline
Hamba \cite{dong2024hamba} & 54.56 & RGB \\
Hybrik \cite{li2021hybrik}      & 2.89 & RGB\\
ArtiBoost \cite{yang2022artiboost}   & 2.53 & RGB\\
HandOccNet   \cite{park2022handoccnet}   & 2.49 & RGB \\
HandVoxNet++ \cite{malik2021handvoxnet++}& 2.46 & Voxel \\
IPNet \cite{ren2023two}      & 1.81  & Depth\\
HandDAGT \cite{cheng2024handdagt}         & 1.81  & Depth \\
K-Fusion \cite{liu2024keypoint}      & 1.79  & RGBD\\
\hline
HandMCM (Ours)         & \textbf{1.71}  & RGBD \\
\hline
\end{tabular}
\caption{Comparison of the proposed method with previous state-of-the-art methods on the HO3D datasets.} 
\label{tab:ho3d}
\end{table}

\subsection{Ablation Study}
\label{sec:abla}
To assess the effectiveness of each component in our proposed method, we 
perform additional ablation studies on different configurations of HandMCM. These experiments 
are mainly conducted using the NYU dataset.
\begin{table}[t]
\small
\centering
\begin{tabular}{c|ccc|c}
\hline
\multirow{3}{*}{ID} & \multicolumn{3}{c|}{Configurations} & Mean \\
\cline{2-4}
& \multirow{2}{*}{SSM Type} & Local & Local & Keypoint\\
&  & Inject  & Filter& Error (mm)\\
\hline
1) & $\times$ & $\times$ & $\times$ &  8.47\\
2) & Transformer & $\times$ & $\times$ &  8.35\\
3) & Standard & $\times$ & $\times$ & 8.51\\
4) & Graph-guided & $\times$ & $\times$ & 8.38\\
\hline
5) & Correspondence & $\times$ & $\times$ & 8.15\\
6) & Correspondence &  $\times$ & $\surd$ & 7.24 \\
7) & Correspondence &  $\surd$ & $\times$ & 7.22 \\
8) & Standard & $\surd$ & $\surd$ & 7.27 \\
9) & Correspondence & $\surd$ & $\surd$ & \textbf{7.06} \\
\hline
\end{tabular}
\caption{Ablations of different configurations of the transformer. All 
ablation models are trained and tested on NYU. } 
\label{tab:abl}

\end{table} 

\begin{table}[t!] \small
\centering
\tabcolsep=3pt
\begin{tabular}{c|c}
\hline
\# of Stacks & Mean Keypoint Error (mm)\\
\hline
1     & 8.83 \\
2   & 7.12 \\
3  & \textbf{7.06} \\
4   & 7.10\\
\hline
\end{tabular}
\caption{Ablations of different depth of Mamba.} 
\label{tab:depth}
\end{table}

\noindent
\textbf{Analysis of Different Proposed Components.}
We conduct an ablation study to evaluate the impact of each proposed component in the key correspondence Mamba block, specifically focusing on the SSM type, using local feature injection, and local feature filtering. The results are summarized in Table \ref{tab:abl}. The specific configurations for the ablation studies are as follows: 1) baseline model without the correspondence Mamba block, 2) Transformer instead of Mamba, 3) standard SSM (S), 4) graph-guided SSM proposed in Hamba \cite{dong2024hamba}, 5) novel SSM with the proposed correspondence modeling, 
6) correspondence SSM added with filtered local features, 7) correspondence SSM with only local token injection, 8) local token injection and filtering both activated on standard SSM, and 9) all proposed components. The baseline model resulted in a mean joint error of 8.47 mm. The ablation using the standard SSM 3) demonstrates an increased error of 8.51 mm, whereas the proposed novel SSM 5) integrated with correspondence modeling plausibly reduces the error by 0.3 mm. This reduction highlights the critical role and effectiveness of the novel SSM. Meanwhile, the proposed correspondence block outperforms the graph-guided SSM 4) proposed in Hamba \cite{dong2024hamba}, revealing the superior correspondence modeling ability of SSM over explicitly applying graphs. Furthermore, incorporating local tokens 6) \& 7) drastically reduced the error to 7.2 mm. Combining local token injection and filtering 9) with Mamba further reduces the error to 7.06 mm, indicating the essential contribution of effective local geometry information. 

\noindent
\textbf{Analysis of Mamba Depth.}
Theoretically, increasing the depth of a network can improve performance by allowing the model to capture more complex features. To investigate this, we progressively expanded the depth of the Mamba block with the aim of identifying its optimal configuration. As shown in Table \ref{tab:depth}, we systematically adjusted the stack depth to assess its effect on model performance. The results indicate that increasing the depth of the Mamba stack generally enhances performance. 

However, the performance gains become negative
beyond three stacks at specifically a 3-stack Mamba.
We thus conclude that a 3-stack configuration is optimal for our model as further depth increases in the transformer stack yield a negative effect.


\section{Conclusion}
This paper introduces HandMCM, an innovative Mamba-based architecture designed for accurate 3D hand pose reconstruction. Experimental results demonstrate that our proposed approach substantially outperforms previous state-of-the-art methods across three challenging datasets. Furthermore, extensive experiments validate the effectiveness of each component proposed in our framework. Nevertheless, despite its effectiveness, our HandMCM still has a limitation in handling scenarios involving bi-manual interacting hands. Future research can focus on extending the model to address this limitation and further enhance its capabilities.

\section{Acknowledgements} 
This research/project is supported by the National Research
Foundation, Singapore, under its NRF-Investigatorship Pro-
gramme (Award ID. NRF-NRFI09-0008).

\end{document}